\documentclass[journal]{IEEEtran}

\usepackage{xcolor,soul,framed} 

\colorlet{shadecolor}{yellow}
\usepackage[pdftex]{graphicx}
\graphicspath{{../pdf/}{../jpeg/}}
\DeclareGraphicsExtensions{.pdf,.jpeg,.png}

\usepackage[cmex10]{amsmath}
\usepackage{array}
\usepackage{mdwmath}
\usepackage{mdwtab}
\usepackage{eqparbox}
\usepackage{url}
\usepackage{tikz}
\usepackage{booktabs}
\usepackage{multirow}
\usepackage{tabularx}
\usepackage{caption}
\begin{document}
\bstctlcite{IEEEexample:BSTcontrol}
    \title{evoML Yellow Paper: Evolutionary AI \\and Optimisation Studio}
  \author{Lingbo Li,
Leslie Kanthan,
Michail Basios,
Fan Wu,\\
Manal Adham,
Vitali Avagyan,
Alexis Butler,
Paul Brookes,
Rafail Giavrimis,\\
Buhong Liu,
Chrystalla Pavlou,
Matthew Truscott,
and~Vardan Voskanyan

\thanks{Lingbo Li,
Leslie Kanthan,
Michail Basios,
Fan Wu,
Manal Adham,
Vitali Avagyan,
Alexis Butler,
Paul Brookes,
Rafail Giavrimis,
Buhong Liu,
Chrystalla Pavlou,
Matthew Truscott,
and~Vardan Voskanyan are with Turing Intelligence Technologies, UK.}
}


\maketitle

\begin{abstract}
Machine learning model development and optimisation can be a rather cumbersome and resource-intensive process.
Custom models are often more difficult to build and deploy, and they require infrastructure and expertise which are often costly to acquire and maintain.
Machine learning product development lifecycle must take into account the need to navigate the difficulties of developing and deploying machine learning models. 
evoML is an AI-powered tool that provides automated functionalities in machine learning model development, optimisation, and model code optimisation. 
Core functionalities of evoML include data cleaning, exploratory analysis, feature analysis and generation, model optimisation, model evaluation, model code optimisation, and model deployment.
Additionally, a key feature of evoML is that it embeds code and model optimisation into the model development process, and includes multi-objective optimisation capabilities.

\end{abstract}

\begin{IEEEkeywords}
code optimisation, multi-objective optimisation, machine learning
\end{IEEEkeywords}

%
\IEEEpeerreviewmaketitle


\section{Motivation and significance}
\label{1_motivation}

Developing rigorous machine learning models is crucial in improving the performance of AI solutions~\cite{brown_machine_2021}. 
However, building custom machine learning models comes with numerous significant challenges, including high demands for time, resources, and specialist expertise. 
Other challenges include the potential for scalability issues, assessing and maintaining data quality, and technical debt in managing code bases \cite{baier_challenges_2019}.

The time required for resolving issues during development is significantly longer for codebases with low code quality compared to those with high code quality~\cite{tornhill2022code}.
In a conventional industrial data science environment, members across different teams (such as data engineers, data scientists, and software engineers) are required to collaborate effectively to take a machine learning model from conceptualisation to deployment. 
Communication across teams can be rather perplexing, adding to the complexities of developing and deploying AI solutions.

These inefficiencies indicate the need for a tool that can remove the burden of developing and optimising machine learning models manually. Automated machine learning (AutoML) is one solution that aims to address some of the concerns of developing and deploying machine learning solutions. AutoML allows users to automate much of the tasks of the model building process such as data cleaning, feature engineering, and model development and evaluation. Automating these tasks significantly reduces the time and costs associated with the machine learning pipeline.

evoML is an automated tool that brings the entire data science cycle onto a single platform. It provides options to generate and deploy machine learning models with a few easy steps, at an expedited rate, with minimal input from data scientists and developers. Contrary to other autoML platforms, a critical element of evoML is its functionalities in multi-objective and code optimisation. 
 
Optimisation can be rather cumbersome, particularly in commercial development environments, as a result of which developers tend to minimise efforts at optimisation or resort to single-objective optimisation.
evoML enables teams to develop and optimise machine learning models with ease, making a strong case of its adoption in commercial software development settings.

\section{Software description}
\label{2_software_desc}

evoML is a software platform that offers a range of capabilities for data preprocessing, feature engineering, model generation, model optimisation, model evaluation, model code optimisation, and model deployment. 
These functionalities can be accessed through a web-based interface or through a workstation client. 
A key component of the platform is the model optimisation feature, which is integrated into the model development process. 
evoML also includes visualisations to aid in the analysis of outputs from each section of the platform. 

\subsection{Glossary of evoML Features and Functionalities}

\textbf{Best Model}: The model that evoML suggests as the one with best performance metrics for a selected task.

\textbf{Data Viewer}: A feature providing users a cross-sectional view of the dataset.

\textbf{Dataset}: The dataset created on evoML using data ingested by users.

\textbf{Deployed Models}: Models that have been deployed to carry out ML-based prediction tasks required by the user.

\textbf{Feature Engineering}: In Feature Engineering, evoML provides automated functionalities to produce more meaningful features from existing ones in the dataset.

\textbf{Features}: This functionality gives features of a given dataset. These features will be used by evoML to generate further insights and visualisations.

\textbf{Green Metrics}: Selecting the Green Metrics feature will include training/prediction carbon emissions and electricity consumption as an objective to be optimised in the model.

\textbf{Machine Learning Task}: evoML offers three machine learning tasks (1) classification, (2) regression, and (3) forecasting. Based on the dataset and the prediction target, the platform sets the machine learning task to one of the above three. Users are also able to change the machine learning task as preferred.

\textbf{ML Models}: ML Models refer to models that have been generated by the platform, including the best model.

\textbf{Relationships}: Relationships provides information and visualisations on correlations observed between different variables of a dataset.

\textbf{Trial}: Term used to refer to the end-to-end model building cycle of evoML. A trial consists of data preprocessing, feature engineering, model building, and model evaluation.

\subsection{Software Architecture}

evoML offers two options for user interaction: a web interface designed for users with limited coding experience, and an evoML client that enables advanced users to integrate the platform into their existing systems. 
This flexible software architecture enables users of all skill levels to access and use evoML's range of machine learning capabilities. 
The web interface provides an easy-to-use, visual interface that allows users to build and optimise models without needing to write code. 
For advanced users who are comfortable with coding, the evoML client provides a code-based interface that can be easily integrated into existing systems and workflows.

The platform guides users through the following phases when moving from conceptualisation to deployment of machine learning models:

First, the \textbf{data preprocessing phase} allows users to upload and view their datasets using the ``dataset" feature.

Next, the \textbf{feature engineering phase} involves selecting and manipulating features from the dataset to create the most effective features for the desired task.
The platform then builds machine learning models for the selected task.

After that, the \textbf{model optimisation phase} uses an iterative process to optimise the developed models and determine the ``best model" for the task. 
The best model, along with relevant metrics and visualisations, is provided to users to help them make informed decisions about their prediction task.

Lastly, \textbf{model code optimisation} and \textbf{model deployment}, evoML provides functionalities for users to optimise model code and deploy models for their use case, which allows users to easily bring their models into a usable state and incorporate them into their workflow or product.

The overall architecture of this process is illustrated in Figure~\ref{fig:evoml-arch}.
For a deeper understanding of the evoML architecture, refer to the documentation available at: \url{https://docs.evoml.ai/}

\begin{figure*}[ht]
    \centering
    \includegraphics[width=0.9\linewidth]{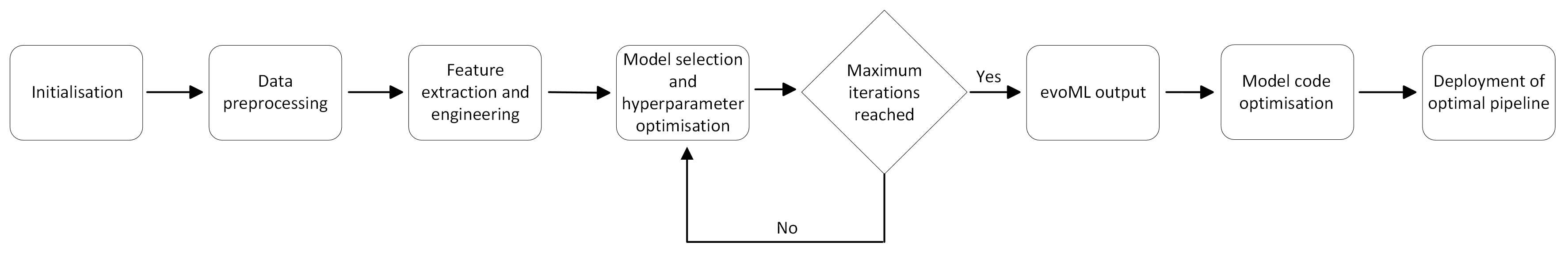}
    \caption{Overview of evoML software architecture}
    \label{fig:evoml-arch}
\end{figure*}

\subsection{Primary Components}

There are two primary components of the evoML platform: (1) Datasets and (2) Trials.

\subsubsection{Datasets}

The datasets component of evoML allows users to upload their data to the platform and perform analysis to identify trends and patterns within the data. 
This feature enables users to gain a better understanding of their data and to inform the development of machine learning models that can effectively extract valuable insights and make accurate predictions.

\textbf{Data upload and preprocessing}:
New data can be uploaded to evoML in one of the following formats:
\begin{enumerate}
    \item Local device: Data files uploaded in the format of CSV, Feather, Parquet, JSON, Avro, and archives.
    \item Database: Data can be sourced from an SQL database, with support for a wide range of platforms including MySQL, PostgreSQL, MongoDB, KDB, and Exasol.
    \item Storage service: Data may be sourced from cloud storage services such as AWS S3, Microsoft Azure, or Minio.
    \item FTP: Data uploaded from an FTP server
\end{enumerate}

Upon uploading data, evoML performs statistical analysis to aid users in exploring the data before building a model. 
This analysis helps users better understand the characteristics and patterns present in the data prior to building the model.

\textbf{Data evaluation}: evoML provides data visualisation tools to help users evaluate the validity of uploaded datasets. 
These visualisations provide a clear and intuitive representation of the data, allowing users to easily identify any potential issues or inconsistencies that may impact the accuracy of their model. 
By thoroughly evaluating their datasets, users can ensure that their models are built on a strong foundation of reliable and relevant data.

\textbf{Feature engineering}: The feature engineering component of evoML has the capability to identify feature correlations and modify and combine features to derive more useful features to be used in the model. 

\begin{figure*}[ht]
    \centering
    \includegraphics[width=0.9\linewidth]{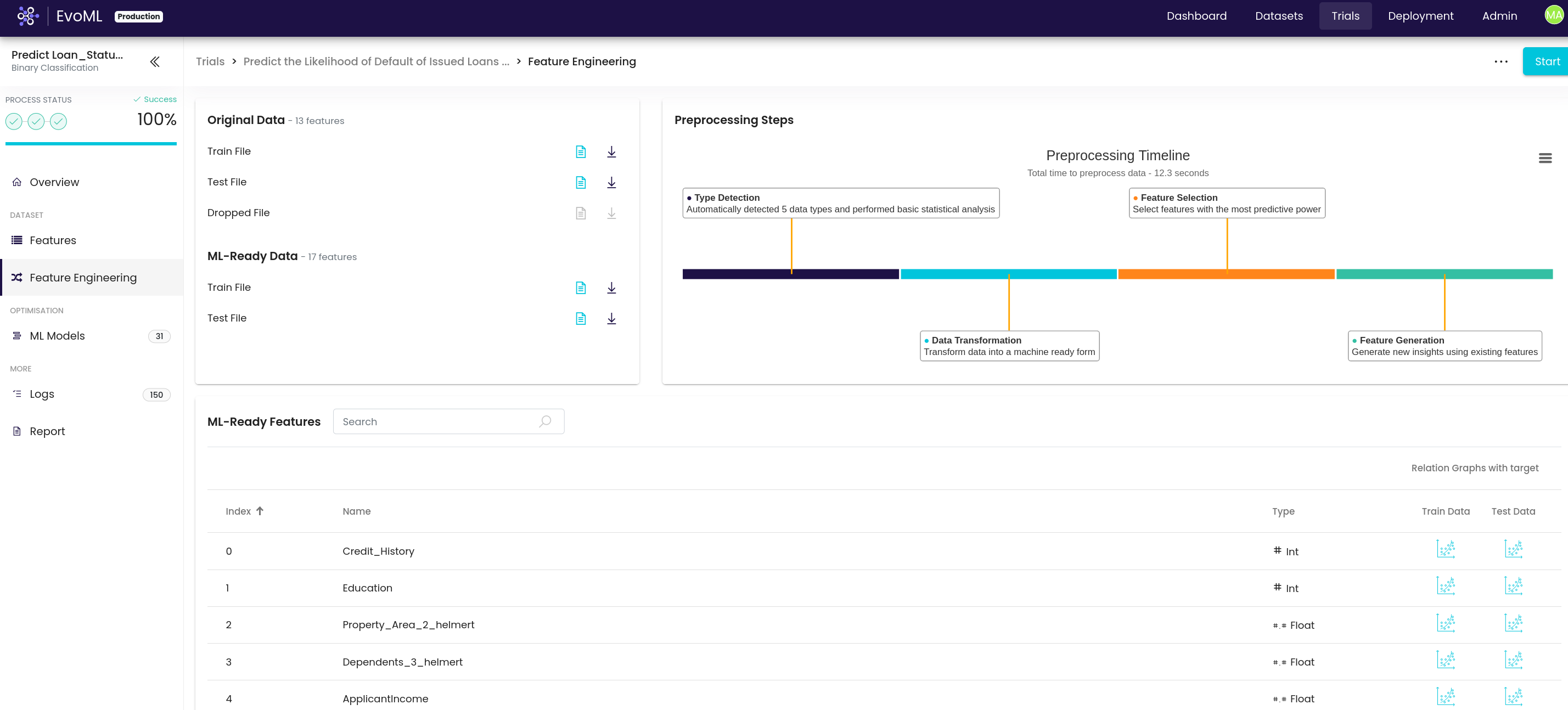}
    \caption{Overview of the feature engineering functionality}
    \label{fig:feature-eng}
\end{figure*}

Figure~\ref{fig:feature-eng} provides an overview of the feature engineering functionality of the platform. 

\subsubsection{Trials}


The Trials component of evoML is a key feature that helps users to develop, evaluate, optimise, and deploy machine learning models. 
This includes tasks such as selecting and preprocessing data, training and evaluating models using various algorithms and hyperparameter configurations, and optimising model performance through techniques such as source code optimisation and internal representation modification. 
Once a model has been developed and optimised, the trials component can then be used to deploy the model for use in production environments. 
Overall, the trials component of evoML provides a comprehensive set of tools and capabilities that enable users to efficiently develop and deploy machine learning models that are optimised for their specific use cases and deliver value to their organisations.

\textbf{Machine learning model creation}: An existing or a newly created dataset can be used to find the optimal machine learning model to carry out a selected machine learning task. 
evoML provides the functionality for a user to select a feature (i.e. a column) to predict, which will provide the basis for the machine learning task.
A range of additional options are provided to refine the scope of the trial. 
Users can also choose to optimise a selected loss function, including green metrics such as energy consumption or carbon emissions. 
For each trial, evoML provides a set of options for feature inclusion, feature selection, and feature generation of the model. 
Users are able to select or omit features from the dataset to be considered in the model development process.
There are also options to customise feature generation options, for instance, by selecting the nature of combinations of variables.

\textbf{Multi-objective optimisation}: As highlighted in the introduction, a valuable feature of evoML is its optimisation capabilities, which enable users to optimise the performance of their machine learning models in various ways. 
During the trial creation process, users can select up to three objectives to optimise. These objectives may include metrics such as training time, prediction time, green metrics, and explainability, among others.
This allows users to tailor their models to the specific requirements and constraints of their use cases. 
For instance, users may want to optimise models for faster training and prediction times in order to reduce the computational resources required for model deployment. 
Alternatively, users may prioritise green metrics, such as energy efficiency or carbon footprint, in order to minimise the environmental impact of their models. 
By providing the ability to optimise models based on a wide range of objectives, evoML enables users to develop and deploy machine learning models that are tailored to their specific needs and constraints.

\textbf{Model explanation and interpretation}: 
evoML provides a range of tools and capabilities for interpreting and evaluating the performance of machine learning models. 
Upon the creation of a model, users can use a variety of visualisations, such as the confusion matrix, ROC curve, precision recall curve, and density plot, to understand the model's behaviour and identify areas for improvement. 
Additionally, evoML provides a range of metrics, such as F1 score, precision, recall, accuracy, and log loss, which are available for train, validation, and test datasets, to help users quantitatively evaluate the performance of the developed models. 
These tools and capabilities enable users to more effectively understand and optimise the performance of their models, helping to ensure that they are delivering accurate and valuable insights.

\textbf{Model Code Optimisation}:
Using evoML, users can identify the optimal model and further enhance the speed and efficiency of the model.
This is achieved through the use of a variety of techniques including lower level source code optimisation techniques~\cite{10.1145/3236024.3236043} and modifications to the internal representation of the model~\cite{nakandala2019compiling}.
This component enables users to improve the speed and efficiency of the model by optimising the way in which it processes data and makes predictions.

\subsection{Supported Machine Learning Models}

Nasteski~\cite{nasteski17} provides an overview of supervised learning models.
Based on those models' effectiveness for certain types of tasks, their popularity and availability of resources to support and maintain them, as well as user demand, evoML consists of a wide variety of machine learning algorithms as well as neural network algorithms for tasks such as classification, regression, and time-series forecasting.
Specifically, the platform includes 46 classification algorithms, 46 regression algorithms, and 6 time-series forecasting algorithms, providing users with a diverse set of tools to choose from when developing machine learning models.

Table~\ref{tab:clf}, Table~\ref{tab:reg}, and Table~\ref{tab:for} list the machine learning classification, regression, and time-series forecasting algorithms that are available within the evoML platform..

\begin{table}[ht]
\caption{Classification models}\label{tab:clf} 
\begin{tabular}{@{}ll@{}}
\toprule
Model type                                                        & Model name                                                                                                                                                                                                                                \\ \midrule\midrule
Bayesian                                                          & \begin{tabular}[c]{@{}l@{}}Gaussian Process, Gaussian Naive Bayes, \\ Bernoulli Naive Bayes, Gaussian Process,\\ Linear Discriminant Analysis, \\ Quadratic Discriminant Analysis\end{tabular}                                            \\ \midrule
Ensemble                                                          & \begin{tabular}[c]{@{}l@{}}Random Forest, Bagging, \\ Extremely Randomized Tree Ensemble, \\ Gradient Boosting, AdaBoost,  \\ CatBoosting, LightGBM\end{tabular}                                                                          \\ \midrule
Gradient                                                          & \begin{tabular}[c]{@{}l@{}}Adaptive Gradient, Coordinate Descent, \\ Fast Iterative Shrinkage/Thresholding, \\ Stochastic Averaged Gradient, \\ Stochastic Averaged Gradient Ascent, \\ Stochastic Variance-reduced Gradient\end{tabular} \\ \midrule
Kernel                                                            & \begin{tabular}[c]{@{}l@{}}Gaussian Process, Label Propagation, \\ Label Spreading, Support Vector Machine, \\ Linear Support Vector Machine, Kernel SVM\end{tabular}                                                                     \\ \midrule
Linear                                                            & \begin{tabular}[c]{@{}l@{}}Logistic Regression, Logistic Regression CV, \\ Ridge, Ridge CV, Perceptron, Passive Aggressive, \\ Stochastic Dual Coordinate Ascent,\\ Stochastic Gradient Descent\end{tabular}                              \\ \midrule
Nearest Neighbors                                                 & \begin{tabular}[c]{@{}l@{}}K-Nearest Neighbors, Nearest Centroid, \\ Radius Neighbors\end{tabular}                                                                                                                                        \\ \midrule
Neural Network                                                     & \begin{tabular}[c]{@{}l@{}}Multilayer Perceptron\\Convolution Neural Network, Recurrent Neural Network\\ Long Short-Term Memory, Gated Recurrent Unit\\Fully Convolutional Network\end{tabular}                                                                                                                                                                                                             \\ \midrule
Semi Supervised                                                   & Label Propagation, Label Spreading                                                                                                                                                                                                        \\ \midrule
\begin{tabular}[c]{@{}l@{}}Support Vector \\ Machine\end{tabular} & \begin{tabular}[c]{@{}l@{}}Support Vector Machine, Kernel SVM,\\ Linear Support Vector Machine\end{tabular}                                                                                                                               \\ \midrule
Tree-based                                                        & \begin{tabular}[c]{@{}l@{}}Random Forest, Gradient Boosting, \\ Decision Tree, CatBoosting, LightGBM\\ Extremely Randomized Tree Ensemble,\\ Extremely Randomized Tree\end{tabular}                                                       \\ \bottomrule
\end{tabular}
\end{table}

\begin{table}[ht]
\caption{Regression models}\label{tab:reg} 
\begin{tabular}{@{}ll@{}}
\toprule
Model type        & Model name                                                                                                                                                                                                                                                                                                                                                                                                \\ \midrule\midrule
Bayesian          & \begin{tabular}[c]{@{}l@{}}Bayesian Ridge,  Gaussian Process,\\ Automatic Relevance Determination\end{tabular}                                                                                                                                                                                                                                                                                            \\ \midrule
Ensemble          & \begin{tabular}[c]{@{}l@{}}Gradient Boosting, Random Forest, \\ AdaBoost, Bagging, CatBoosting, \\ LightGBM\end{tabular}                                                                                                                                                                                                                                                                                  \\ \midrule
Gradient          & \begin{tabular}[c]{@{}l@{}}Adaptive Gradient, Coordinate Descent, \\ Fast Iterative Shrinkage/Thresholding, \\ Stochastic Averaged Gradient, \\ Stochastic Averaged Gradient Ascent\end{tabular}                                                                                                                                                                                                          \\ \midrule
Kernel            & Kernel Ridge, Gaussian Process                                                                                                                                                                                                                                                                                                                                                                            \\ \midrule
Linear            & \begin{tabular}[c]{@{}l@{}}Linear Regression, Ridge, Ridge CV, \\ Lasso, Lasso CV, Elastic Net, Elastic Net CV, \\ Least Angle, Lasso Lars, Bayesian Ridge, \\ Automatic Relevance Determination, \\ Stochastic Gradient Descent, Passive Aggressive, \\ Random Sample Consensus, Huber, Theil-Sen,\\ Partial Least Squares, Stochastic Dual Coordinate Ascent\\ Orthogonal Matching Pursuit\end{tabular} \\ \midrule
Nearest Neighbors & K-Nearest Neighbors, Radius Neighbors                                                                                                                                                                                                                                                                                                                                                                     \\ \midrule
Neural Network     & \begin{tabular}[c]{@{}l@{}}Multilayer Perceptron\\ Convolution Neural Network, Recurrent Neural Network\\Long Short-Term Memory, Gated Recurrent Unit\\Fully Convolutional Network\end{tabular}                                                                                                                                                                                                                                                                                                                  \\ \midrule
Tree-based        & \begin{tabular}[c]{@{}l@{}}Decision Tree, Extremely Randomized Tree, \\ Gradient Boosting, Random Forest, \\ CatBoosting, LightGBM\end{tabular}                                                                                                                                                                                                                                                           \\ \bottomrule
\end{tabular}
\end{table}

\begin{table}[ht]
\caption{Forecasting models}\label{tab:for} 
\begin{tabular}{@{}ll@{}}
\toprule
Forecasting models & \begin{tabular}[c]{@{}l@{}}Auto ARIMA Forecaster, Auto ETS,  \\ Local Global Trend Forecaster, Naive Forecaster, \\ Prophet Forecaster, Damped Local Trend Forecaster\end{tabular} \\ \bottomrule
\end{tabular}
\end{table}

\subsection{Generative AI and code optimisation}

evoML is centred around the fundamentals of generative AI and nature-inspired optimisation (Reinforcement Learning~\cite{kaelbling1996reinforcement}, Evolutionary Algorithm~\cite{zhou2011multiobjective}, Bayesian Optimisation\cite{garnett_bayesoptbook_2023} etc.).
With data as input, evoML generates optimised machine learning models. 
Code optimisation allows evoML to further optimise models at code level, and users are able to go through model code to get a clear sense of the model's prediction process. 
Giavrimis et al.~\cite{9678650} conducted a study to assess the impact of optimising the codebase of the mlpack machine learning library on its performance. 
They found that through code optimisation, the library was able to achieve a $27.9\%$ reduction in execution time and a $2.7\%$ reduction in memory usage while maintaining the library's predictive capabilities.

\section{Illustrative Examples}
\label{3_examples}

\textbf{Customer churn prediction with evoML}

\begin{figure*}[ht]
    \centering
    \includegraphics[width=0.9\linewidth]{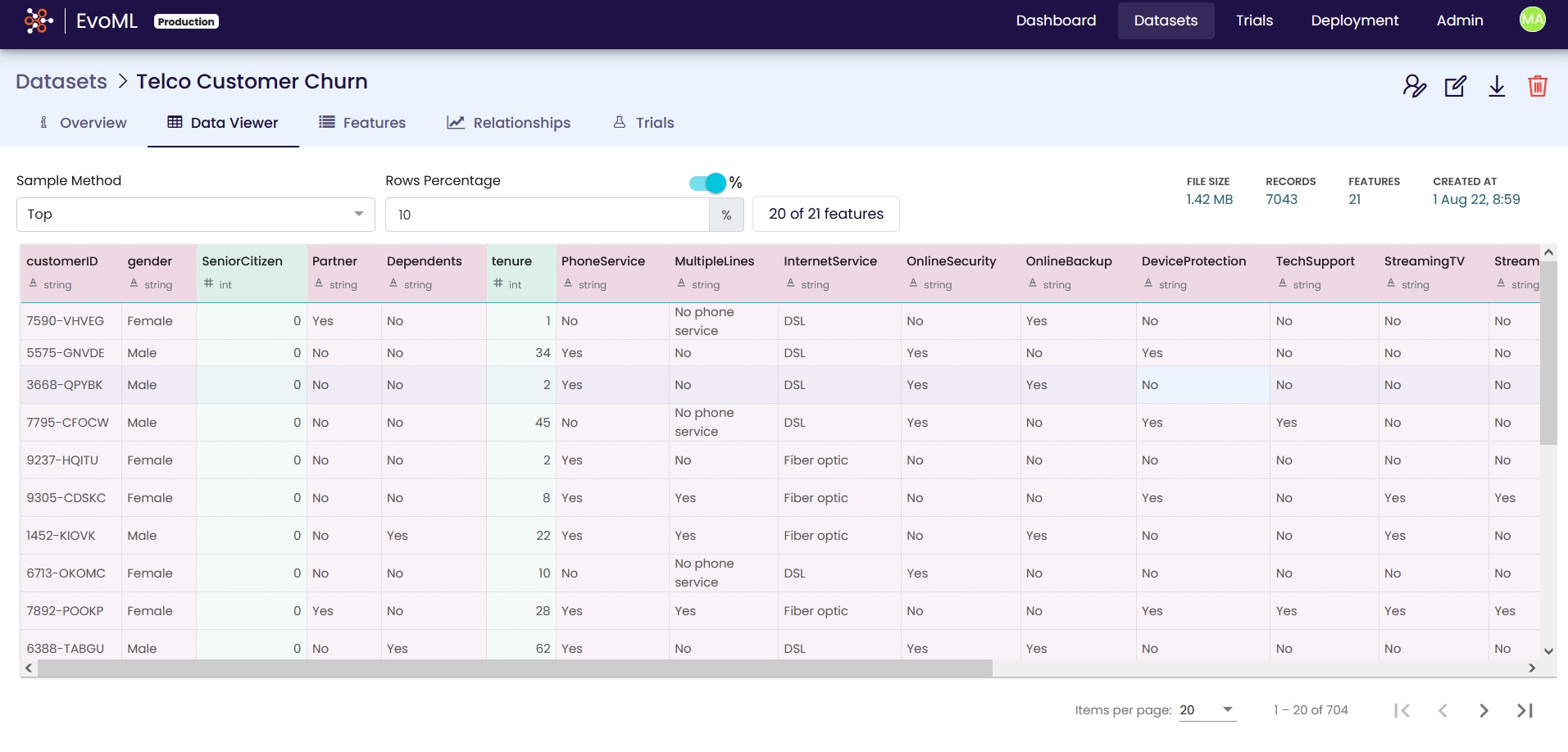}
    \caption{Overview of churn dataset}
    \label{fig:churn-1}
\end{figure*}

This example considers a customer churn dataset taken from Kaggle\footnote{available at: https://www.kaggle.com/datasets/blastchar/telco-customer-churn)}. 
The dataset contains information on a fictional telecommunications company that offers home phone and internet services to customers. 
It captures information of 7,043 customers across 21 columns. 
These columns, such as ``gender," become features in the model.
Figure~\ref{fig:churn-1} gives a snapshot of the dataset uploaded to evoML. 

The example will use the above dataset to build a model that can predict whether a given customer is likely to churn or not.
This is a classification task, with \textbf{churn} being the target feature.

To develop the model, users begin by creating a ``trial" on evoML and selecting the appropriate dataset and task. 
The platform then uses a range of techniques to develop and optimise a model that is best suited for the task at hand. 

The feature engineering overview allows users to see the changes that evoML has applied to the existing features in the dataset.

Upon the completion of the ``trial", evoML gives the best model for the churn prediction task, along with additional information to evaluate its performance. 

The platform also includes a deployment option that exposes the best model and makes it available for instant churn predictions.

\section{Impact}

\label{4_impact}

evoML is a unique platform that integrates the entire data science cycle into a single, cohesive environment. 
While other automated machine learning (autoML) platforms offer automated model building capabilities, none of them currently provide the ability to optimise the models within the same automated process. 
evoML stands out as the only platform that offers automated model and code optimisation as a core feature, enabling users to optimise the performance of their machine learning models and artificial intelligence solutions in a way that is consistent with net-zero impact goals. 

Additionally, evoML's multi-objective optimisation functionality allows users to optimise models over multiple hyperparameters, further improving performance while also considering the environmental impact of the model's deployment. 
These optimisation capabilities make it easy for data scientists and developers to incorporate optimisation into their workflows in a way that is aligned with net-zero impact objectives, ultimately leading to better overall performance of machine learning models.

\section{Conclusions}
\label{5_conclusion}

evoML is a comprehensive automated machine learning platform that provides a range of functionalities for data wrangling, feature engineering, model development, model evaluation, model code optimisation, and model deployment. 
These capabilities can be accessed through the evoML user interface as a no-code option or through the evoML client for users with coding experience. 
A key feature of evoML is its multi-objective optimisation capability, which enables users to optimise models based on multiple criteria. 
This is a unique feature that is not currently offered by other autoML platforms. 
The inclusion of multi-objective optimisation in evoML helps to overcome the challenges and resource constraints that can often hinder manual code and optimisation efforts. 
As a result, developers can more easily implement optimisation features and build high-performing machine learning models without putting in significant time and effort. 
Overall, evoML is a valuable tool for anyone looking to automate the machine learning model development and optimisation process.


%





\ifCLASSOPTIONcaptionsoff
  \newpage
\fi





\bibliographystyle{IEEEtran}
\bibliography{IEEEabrv.bib}





\vfill


\end{document}